\documentclass[cameraready]{Interspeech}
\usepackage{algorithm}
\usepackage{algpseudocode}
\usepackage{subcaption}
\usepackage{amsmath,amssymb}
\usepackage[most]{tcolorbox}
\usepackage{xcolor}

\title{Towards Dys-XAI: Influence-Based Explanations for \\ Dysarthria Severity Assessment}
\author[affiliation={1}]{Xiaoliang}{Wu}
\author[affiliation={2}]{Qiyang}{Sun}
\author[affiliation={2}]{Yupei}{Li}
\author[affiliation={3}]{Erfan}{Loweimi}
\author[affiliation={1}]{Jennifer}{Williams}
\author[affiliation={4}]{Zhengjun}{Yue}

\address{
    $^1$ University of Southampton, United Kingdom \\
    $^2$ Imperial College London, United Kingdom \\
    $^3$ University of Edinburgh, United Kingdom \\
    $^4$ King's College London, United Kingdom
}

\email{xiaoliang.wu@soton.ac.uk, zhengjun.1.yue@kcl.ac.uk}


\usepackage{comment}

\begin{document}

\maketitle

\begin{abstract}


Dysarthria severity assessment is essential for therapy planning and longitudinal monitoring, yet manual perceptual rating is time-consuming and variable across clinicians. Although deep learning models achieve strong performance, their black-box nature limits clinical adoption. Existing speech explainability methods typically provide acoustic feature importance scores that are difficult for end-users to interpret. We propose an influence-based, instance-level explainability framework that explains each decision through supportive and competing training samples.
Using gradient-based influence approximations, we compute per-utterance influence scores to identify supportive and competing training samples for each prediction. Controlled deletion experiments (5–20\%) validate the explanations, showing that removing highly influential samples systematically shifts predictions. This approach provides auditable explanations by linking decisions to perceptible reference cases.



%
\end{abstract}

\section{Introduction}

Dysarthria severity assessment aims to quantify the degree of speech motor impairment in individuals, playing a critical role in clinical therapy planning, rehabilitation monitoring, and longitudinal disease tracking \cite{yorkston1996comprehensibility,duffy2012motor}. In current clinical practice, severity is commonly judged via auditory–perceptual rating protocols, which are time-consuming and subject to inter-rater variability \cite{kent1999acoustic,stipancic2021you}. 
Automatic severity assessment, therefore, offers the prospect of fast, standardised, and repeatable ratings, reducing clinician workload and enabling more frequent monitoring.
Recent deep learning-based methods \cite{yue23_interspeech,anuprabha2025dysarthric,deshpande2025sand,yeo2026multilingual} have achieved performance comparable to expert ratings on multiple public dysarthric speech datasets \cite{rudzicz2012torgo,kim2008dysarthric,sun2023cdsd}, demonstrating their potential for clinical decision support.

However, assessment performance alone is insufficient for real-world deployment.
Model explainability, i.e., the ability to explain \textit{``why a particular severity level was assigned''}, is crucial for clinical adoption and user trust \cite{xu6019779evaluating,chazette2021exploring}. In practice, both clinicians and patients are often hesitant to act on black-box models that lack transparent, clinically meaningful justification, especially when outputs may influence therapy planning and prognostic discussions \cite{xu2025evaluating}.
%
Explainability in this setting is therefore not only a tool for researchers to inspect model behavior, but also provides evidence that is understandable and reviewable by end users (clinicians and, where appropriate, patients), supporting verification and communication. 

Current explainability research in speech mostly employs post-hoc feature attribution \cite{wu2023,wu2024explainableattributebasedspeakerverification}, including time–frequency saliency approaches (e.g., Shapley Additive Explanations (SHAP) \cite{shap-ig,shap-2,shap-3}, Local Interpretable Model-agnostic Explanations(LIME) \cite{lime-1,lime-2,wu2023trustexplainableaimethods}, attention-based visualisation \cite{attention1,attention2}), feature-importance analyses over predefined acoustic feature sets such as OpenSMILE descriptors \cite{opensmile}, and model layer behaviour \cite{yue2025probing}. These methods typically produce visual or numerical importance representations, e.g., heatmaps over spectrograms
or ranked feature-importance scores, that indicate which regions or descriptors contributed most to a model's output for a given utterance.
While informative about local model sensitivity, their practical interpretive value in clinical severity assessment is often limited. 
First, these explanations are local and utterance-level. Severity labels are inherently ordinal: adjacent categories (e.g., mild vs. moderate) reflect incremental differences along an impairment continuum rather than sharply separable nominal classes. A clinically useful explanation should therefore expose relational evidence—how an utterance compares to neighbouring severity levels—which is not directly conveyed by within-utterance saliency or descriptor rankings.
Second, clinicians often reason in perceptual constructs, e.g., consonant precision, vowel centralisation, abnormal prosody, slow rate, or breathiness \cite{nagle2018perceived}, making it non-trivial to translate spectrogram heatmaps or numerical importance of engineered descriptors into clinically meaningful justifications.

In this work, we shift the explanation perspective from acoustic features to case-based training instances. Instead of asking ``\textit{Which time–frequency regions or acoustic features were important?}'', we ask: ``\textit{Which training samples exert positive influence on this prediction, and which exert opposing influence?}''
This case-based view makes the ordinal structure explicit, because influence can be inspected and aggregated by severity label, rather than only by internal feature contributions.
Importantly, because training instances are audio recordings, the explanations can be directly presented as perceptible reference examples, and aligning naturally with how clinicians justify judgements through comparison with prototypical examples.

In this paper, to operationalise this perspective, we construct a structured influence-based explainability framework built on gradient-based approximation of training-sample influence \cite{tracin}. 
 We innovate the influence methods in previous work \cite{koh2020influence,tracin} to obtain a per-test-utterance influence score over the entire training set, and perform structured aggregation and analysis at two complementary levels: (i) \textbf{class-level} aggregation groups influence scores by severity label to quantify supportive vs. competing contributions across the ordinal scale, and (ii) \textbf{model-level} aggregation aggregates influence patterns across test cases to characterize the model's global reliance on different severity levels.
Finally, we evaluate the explanatory validity with \textbf{controlled deletion experiments}: By selectively removing highly influential (or strongly opposing) training cases and measuring the systematic changes in model outputs, we test whether the proposed influence-based explanations correspond to causal, prediction-relevant evidence rather than correlational artifacts. 


\section{Methodology}

\subsection{Task formulation}
We follow the severity annotation provided in TORGO \cite{rudzicz2012torgo} and formulate dysarthria severity assessment as a 4-class ordinal classification task. Given an utterance $x$, the label is $y \in \{0, 1, 2, 3\}$, corresponding to \{typical, mild, moderate, and severe\} dysarthria (with moderate-to-severe merged into severe).
The training set is $\mathcal{D}_{\text{train}} = \{(x_i, y_i)\}_{i=1}^{N}$, where $i$ represent the index of the sample. The model being explained (a 4-class classifier in our case) $f_\theta$ is trained by minimising cross-entropy loss:

\begin{equation}
\mathcal{L}(\theta) = \frac{1}{N}\sum_{i=1}^{N} \ell(f_\theta(x_i), y_i),
\end{equation}
where $\ell(\cdot, \cdot)$ denotes per-sample loss.

Our goal is to explain a prediction on test sample $(x_t, y_t)$ by quantifying which training samples support it, which oppose it and how such influence distributes across severity levels.

\subsection{Instance-level influence computation}
To quantify how an individual training sample affects model prediction on a test sample, we adopt a gradient-based influence estimation method~\cite{tracin}. This method measures influence by tracking gradient alignment throughout the training trajectory. 
%
Given checkpoints $\{\theta_1, \ldots, \theta_M\}$ saved during training, the influence of training sample $x_i$ on model's prediction on test sample $x_t$ throughout training is defined as the cumulative gradient inner product over checkpoints:
In our experiments, we save one checkpoint after each training epoch, giving $M=30$ checkpoints for each fold.

\begin{equation}
I_{i \to t} = \sum_{m=1}^{M} \nabla_\theta \ell(f_{\theta_m}(x_t), y_t)^\top \nabla_\theta \ell(f_{\theta_m}(x_i), y_i).
\end{equation}

 
At each checkpoint $m$, the gradient inner product measures directional alignment:
\begin{itemize}
    \item \textbf{Positive inner product}: gradients point in similar directions. Training on $x_i$ would reduce loss on $x_t$ at this point---$x_i$ provides supportive influence.
    \item \textbf{Negative inner product}: gradients point in opposite directions. Training on $x_i$ would increase loss on $x_t$---$x_i$ exerts counteracting influence.
\end{itemize}


For each test sample $x_t$, computing $\mathbf{I}_t = [I_{1 \to t}, I_{2 \to t}, \ldots, I_{N \to t}]$ yields an influence ranking over the training set, enabling us to identify the most supportive and most counteracting training samples for that specific test case.


\subsection{Class-level influence aggregation}
Instance-level influences identify specific training utterances that affected each prediction. To reveal systematic patterns, we aggregate by the test samples' true severity labels.

For each severity level $c \in \{0, 1, 2, 3\}$, we collect all test samples with true label $c$:

\begin{equation}
\mathcal{T}_c = \{x_t \in \mathcal{D}_{\text{test}} : y_t = c\}.
\end{equation}

For each training sample $x_i$, we compute its \textbf{average influence} across all test samples in $\mathcal{T}_c$:

\begin{equation}
\mathcal{I}_{i}^{(c)} = \frac{1}{|\mathcal{T}_c|} \sum_{x_t \in \mathcal{T}_c} I_{i \to t}.
\label{eq:class_level influence}
\end{equation}

This produces a \textbf{class-aggregated influence vector} $\boldsymbol{\mathcal{I}}^{(c)} = [\mathcal{I}_{1}^{(c)}, \mathcal{I}_{2}^{(c)}, \ldots, \mathcal{I}_{N}^{(c)}]$, where each entry measures the average influence of training sample $x_i$ on test samples at severity level $c$. Averaging accounts for varying test set sizes across severity levels, making influence scores comparable.
Training samples can be ranked by $\mathcal{I}_{i}^{(c)}$ to identify the most influential cases for each severity group. Samples with high positive $\mathcal{I}_{i}^{(c)}$ provided strong support for predictions at level $c$, while those with large negative $\mathcal{I}_{i}^{(c)}$ opposed such predictions.

To analyse cross-level influence patterns, we further decompose by training labels. For each pair of severity levels $(c, c')$, we compute the \textbf{average influence} from training samples labeled $c'$ on test samples labeled $c$:
\begin{equation}
S_{c \leftarrow c'} = \frac{1}{|\{i : y_i = c'\}|} \sum_{i : y_i = c'} \mathcal{I}_{i}^{(c)}.
\label{eq:matrix_S}
\end{equation}

The matrix $\mathbf{S} = [S_{c \leftarrow c'}]_{c, c' \in \{0,1,2,3\}}$ forms a $4 \times 4$ \textbf{class-level influence matrix}, where entry $S_{c \leftarrow c'}$ represents the average per-sample influence from training level $c'$ to test level $c$.
Normalizing by the number of training samples at level $c'$ ensures that $S_{c \leftarrow c'}$ values are comparable across severity levels, regardless of class imbalance in either the training or test sets. 


\subsection{Model-level influence analysis}
To quantify the model's sensitivity to ordinal structure, we aggregate influence 
by ordinal distance $d = |c - c'|$:

\textbf{Ordinal sensitivity assessment.} The class-level influence matrix $\mathbf{S}$ captures how training samples from each severity level contribute to predictions on each test severity group. Severity labels are ordered: adjacent levels (e.g., mild vs.\ moderate) represent incremental distinctions, while distant levels (e.g., typical vs.\ severe) represent coarse contrasts. A model that respects this ordinal structure should rely more heavily on nearby training samples than distant ones.

We define \textbf{ordinal distance} between training label $c'$ and test label $c$ as $d(c', c) = |c' - c|$. For each distance $d \in \{0, 1, 2, 3\}$, we compute the average matrix entry at that distance. For example, $\bar{S}(0)$ averages the diagonal entries $\{S_{0 \leftarrow 0}, S_{1 \leftarrow 1}, S_{2 \leftarrow 2}, S_{3 \leftarrow 3}\}$ (same-level support), while $\bar{S}(1)$ averages adjacent entries $\{S_{0 \leftarrow 1}, S_{1 \leftarrow 0}, S_{1 \leftarrow 2}, S_{2 \leftarrow 1}, S_{2 \leftarrow 3}, S_{3 \leftarrow 2}\}$ (adjacent-level support). Formally:
\begin{equation}
\bar{S}(d) = \frac{1}{|\{(c, c') : |c - c'| = d\}|} \sum_{c, c' : |c - c'| = d} S_{c \leftarrow c'}.
\label{eq:matrix_barS}
\end{equation}

If the model exhibits ordinal sensitivity, $\bar{S}(d)$ should decrease monotonically: the samples at the same level ($d=0$) provide the strongest support, adjacent samples ($d=1$) moderate support, and distant samples ($d \geq 2$) weak or negative support. Deviations reveal how the model encodes severity structure: if $\bar{S}(2) > \bar{S}(1)$, the model relies on extreme contrasts rather than fine-grained distinctions; if $\bar{S}(d)$ is flat, this suggests that the model treats severity as nominal rather than ordinal.

\section{Experiments}

\begin{figure}[ht!]
\centering
\includegraphics[width=0.9\columnwidth]{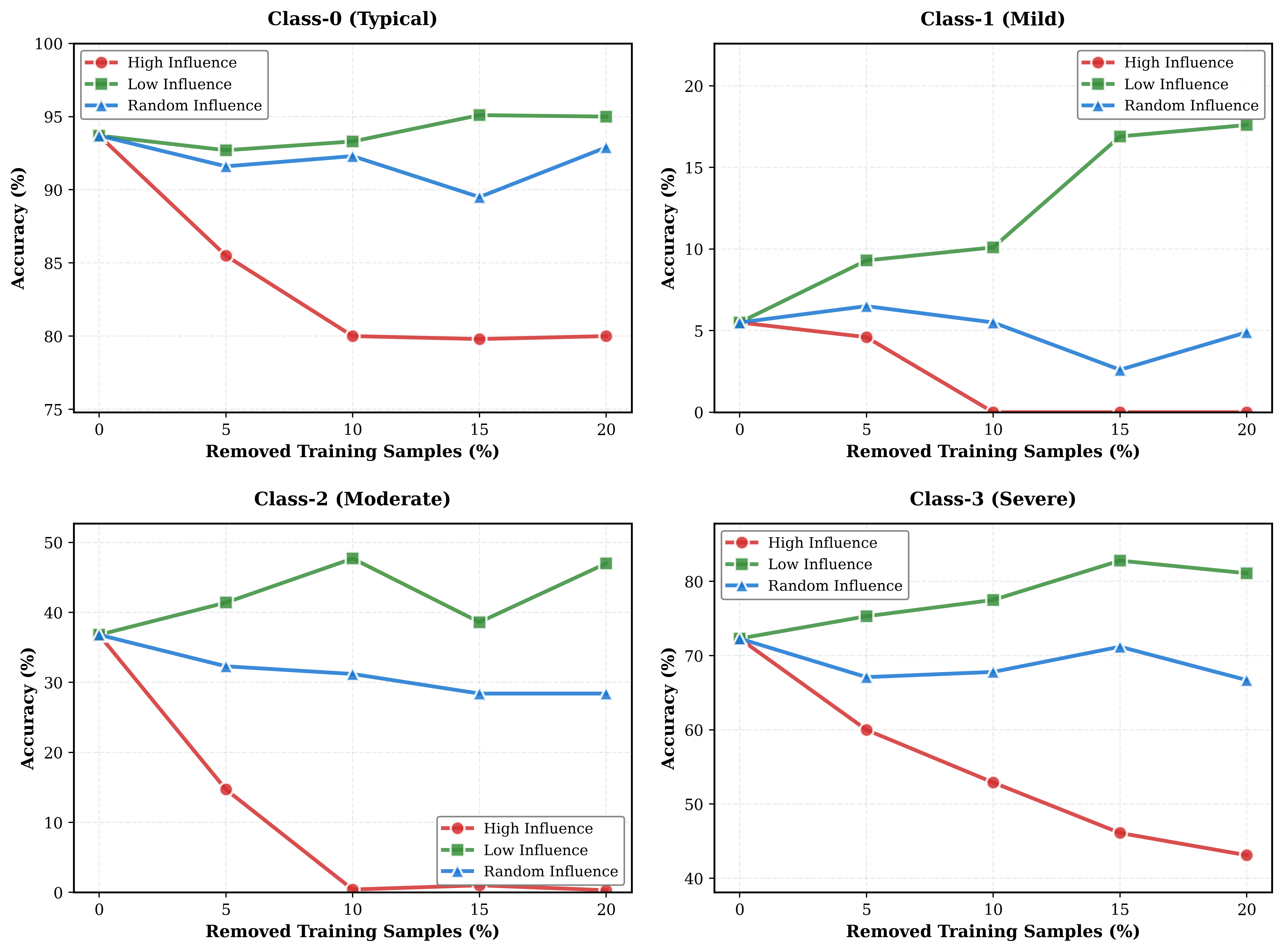}
\caption{%
Validation from controlled deletion strategies indicate 
performance changes across severity levels. \textcolor{red}{\LARGE$\bullet$}: high influence removal. \textcolor{green}{\small$\blacksquare$}: low influence removal. \textcolor{blue}{\small$\blacktriangle$}: random removal.
}
\vspace{-10pt}
\label{fig:deletion}
\end{figure}

We conduct experiments on the TORGO \cite{rudzicz2012torgo}, a widely used English dysarthric dataset, which contains approximately 21 hours of recordings from 15 speakers. TORGO includes 8 speakers with dysarthria (7.3 hours) and 7 typical speakers (13.7 hours). The speech material comprises both isolated words and sentences.
Severity labels are provided at the speaker level. In total, TORGO contains 17,587 utterances.

We use stratified $K$-fold cross-validation \cite{Kohavi1995,scikit-learn} with group-wise splitting by speaker.
Stratification is performed with respect to the speaker-level severity labels to keep class proportions as balanced as possible across folds.
In each run, one fold is used for testing, and the remaining folds are used for training and validation. 

For the base classifier model, we extract 80-dimensional filterbank (FBank) features and train a linear classifier with a single fully connected layer followed by a softmax.
Each classifier was trained for 40 epochs per fold using the AdamW optimizer \cite{zhou2024towards} with a learning rate of 3e-4 and a batch size of 32. For the controlled deletion experiments, we use the same model architecture and training protocol on the modified training sets.

\section{Results and discussion}


We present the outcome of evaluating our explanation framework quantitatively and qualitatively. 
Our results summarise the faithfulness of our proposed influence scores via a set of controlled deletion experiments. Finally, we present our analysis of cross-severity influence patterns and qualitative case studies.

\subsection{Validation of influence-based framework}
A critical component of our evaluation is to first validate whether computed influence scores meaningfully capture training sample importance. This is important because influence-based explanations are only useful if high-influence samples genuinely drive predictions. We employ controlled deletion experiments to test this directly. For each test level $c$, we rank training samples by their class-aggregated influence $\mathcal{I}_i^{(c)}$ (Eq.~\ref{eq:class_level influence}) and remove samples under three strategies. Results shown in Figure~\ref{fig:deletion} indicate that the three strategies produce clearly 
divergent trajectories, confirming that influence scores are capturing systematic 
differences in sample importance. 


\begin{figure}[ht!]
\centering
\includegraphics[width=0.8\columnwidth]{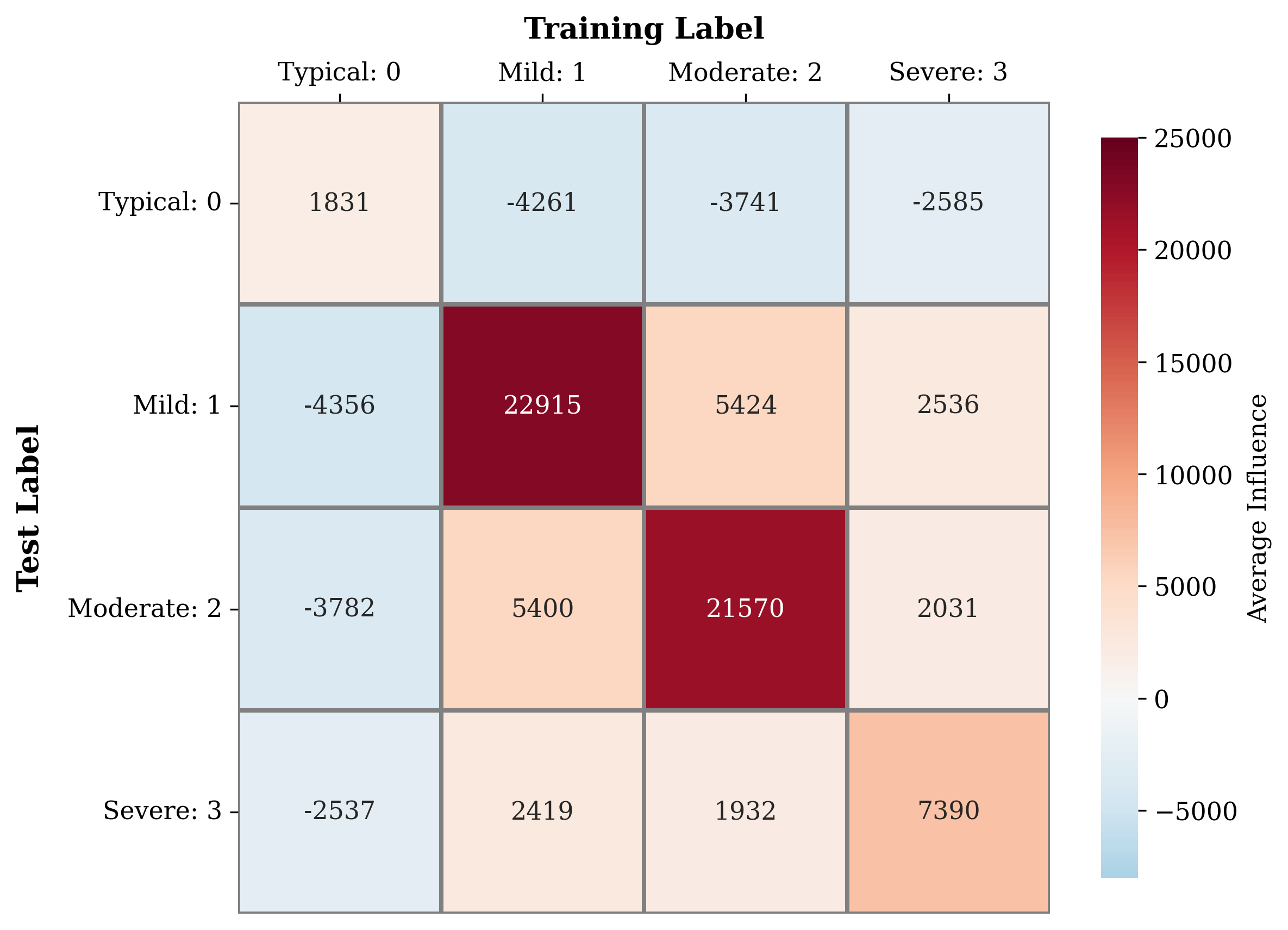}
\caption{ Class-level influence matrix $\mathbf{S}$. Rows = test labels, columns = training labels. Red (dark) positive (supporting predictions). Blue (light) negative (opposing predictions). }
\vspace{-10pt}
\label{fig:heatmap}
\end{figure}

\textbf{Strategy: Random.} This strategy removes $k\%$ samples uniformly at random, serving as the baseline control. \textbf{Random removal shows minimal change}. The delta values across all levels ($|\Delta| < 2\%$), confirm that observed effects stem from influence-guided selection, not dataset size reduction.

\textbf{Strategy: High Influence.} Removing the top-$k\%$ samples with highest $\mathcal{I}_i^{(c)}$ can indicate whether influence scores are valid, because performance on test level $c$ should degrade more than the random baseline. \textbf{High influence removal degrades performance substantially.} Class-2 
(Moderate) drops from 36.8\% to 0.3\% at 20\% removal ($\Delta = -36.5$ points), and Class-3 (Severe) falls from 72.3\% to 43.1\% ($\Delta = -29.2$ points). Even Class-0 (Typical), with 93.7\% baseline accuracy, loses 13.7 points. Class-1 (Mild) collapses from 5.5\% to near-zero (0.3\%). Consistent degradation across severity levels validates that influence ranking identifies critical training samples.

\textbf{Strategy: Low Influence.} After removing the bottom-$k\%$ samples with lowest $\mathcal{I}_i^{(c)}$, this is expected to outperform the baseline by eliminating noisy or mislabeled samples. We retrain the model on each reduced training set and evaluate classification accuracy on test samples at severity level $c$ for removal percentages $k \in \{0, 5, 10, 15, 20\}$. \textbf{Low influence removal improves performance.} Class-2 increases from 36.8\% to 47.0\% (+10.2 points), Class-3 from 72.3\% to 81.1\% (+8.8 points), and Class-1 more than triples from 5.5\% to 17.6\% (+12.1 points). These gains indicate that low-influence samples include noise or mislabeled cases that degrade model quality. 
The largest improvements occur for the lowest-baseline classes (Class-1 and Class-2), suggesting influence-based filtering is most valuable for ambiguous severity levels.


These results validate our framework. High-influence removal degrades performance, confirming critical samples are identified. Low-influence removal improves performance, confirming noisy samples are detected. After validating that influence scores reliably capture training sample \textit{importance}, we can now examine how their influence distributes \textit{across severity levels}, addressed in the following section.

\subsection{Cross-severity influence patterns}


\subsubsection{Class-level influence matrix}
%
In Figure~\ref{fig:heatmap}, we visualize the class-level influence matrix $\mathbf{S}$ (Eq.\ref{eq:matrix_S}). Entry $S_{c \leftarrow c'}$ represents the average influence from training samples at severity level $c'$ on test samples at level $c$. From this $4 \times 4$ matrix visualisation, clear structural patterns are revealed. 
The matrix shows block-diagonal structure. Diagonal entries dominate their rows, with mild-to-mild ($S_{1 \leftarrow 1} = 22915$) and moderate-to-moderate ($S_{2 \leftarrow 2} = 21570$) showing the strongest influence. The structure confirms that predictions rely primarily on same-level training evidence. Typical-to-typical influence ($S_{0 \leftarrow 0} = 1831$) is comparatively lower, and this is because typical samples are prototypical and require less discriminative 
evidence to establish their category.

Beyond the diagonal, the structure reveals a critical finding. Typical test samples receive positive influence only from typical training samples and all dysarthric levels exert negative influence. The strongest opposition comes from mild samples ($S_{0 \leftarrow 1} = -4261$). This pattern is reasonable because mild dysarthria marks the boundary between typical and atypical speech, providing discriminative anchors that prevent false-typical predictions.

Within dysarthric levels, the pattern inverts. Mild, moderate, and severe test samples draw positive influence from each other's training samples. Moderate, for instance, receive support from mild ($S_{2 \leftarrow 1} = 5400$) and severe ($S_{2 \leftarrow 3} = 2031$) samples beyond moderate samples themselves. 
This mutual support indicates that the model recognises shared acoustic features across dysarthric speech, regardless of severity.

The matrix thus encodes a two-tier structure: a binary distinction separating typical from dysarthric speech, layered with ordinal grading among dysarthric severities. Typical stands isolated; dysarthric levels form a cluster with internal gradation. 

\subsubsection{Model-level ordinal sensitivity}
\begin{figure}[t]
\centering
\includegraphics[width=0.8\columnwidth]{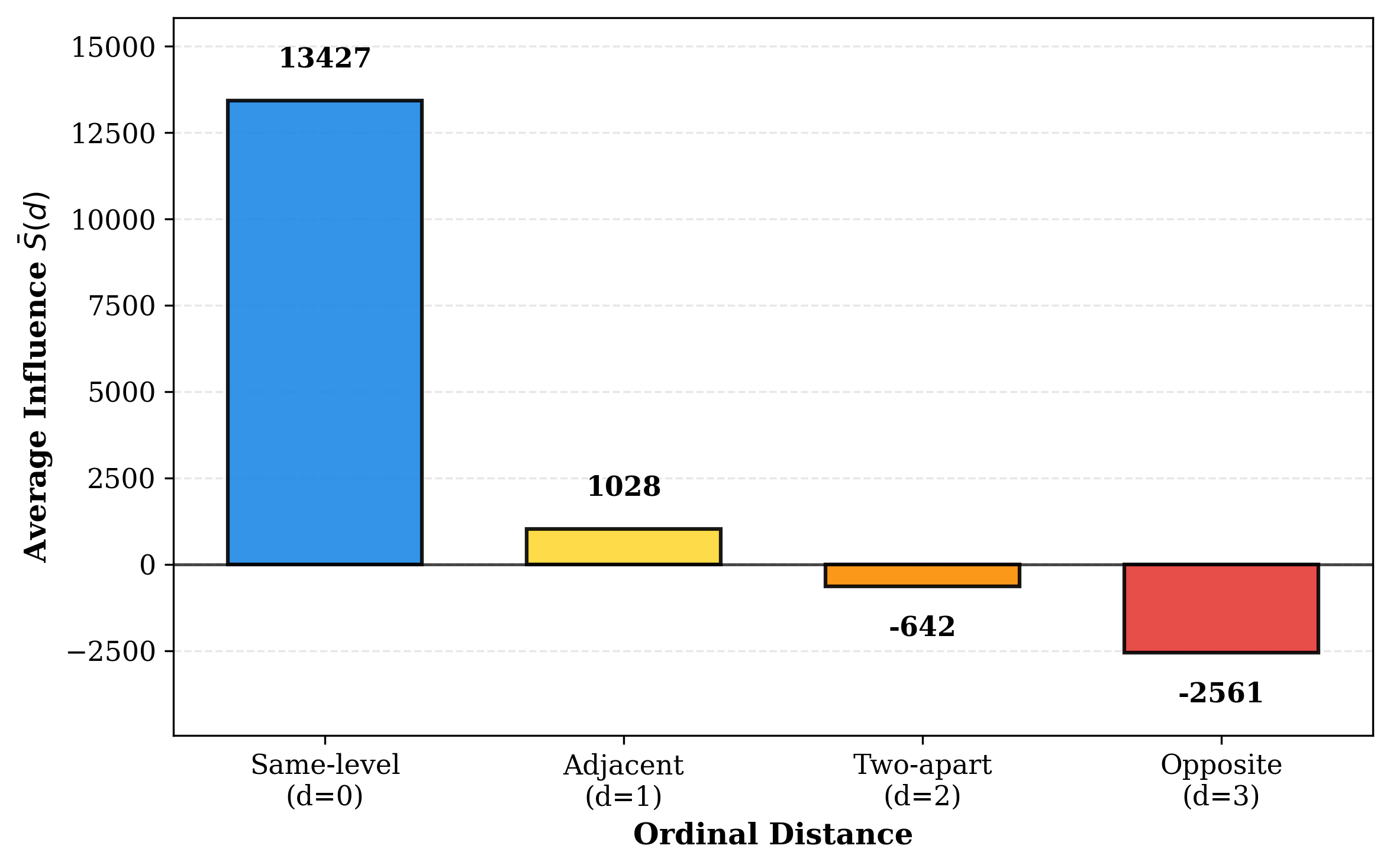}
\caption{ Ordinal sensitivity analysis. Average influence $\bar{S}(d)$ decreases monotonically with ordinal distance $d$.}
\vspace{-10pt}
\label{fig:ordinal}
\end{figure}

%
To quantify the model's sensitivity to ordinal structure, we aggregate influence by ordinal distance $d = |c - c'|$ (Eq.\ref{eq:matrix_barS}). Figure~\ref{fig:ordinal} shows the average influence $\bar{S}(d)$ for each distance level.

We observe a monotonic decay: $\bar{S}(0) = 13427$, $\bar{S}(1) = 1028$, $\bar{S}(2) = -642$, $\bar{S}(3) = -2561$. Same-level support ($d=0$) is 13$\times$ larger than adjacent-level influence ($d=1$), and influence becomes negative for $d>=2$. This graded transition from supportive to opposing influence as distance increases is consistent with an ordinal-sensitive model rather than nominal categories: neighbouring severities provide supportive evidence, whereas distant severities exert counteracting influence. 



\subsection{Case study: instance-based vs. feature attribution}
\begin{figure}[ht]
\centering
\begin{subfigure}{0.45\linewidth}
  \includegraphics[width=\linewidth]{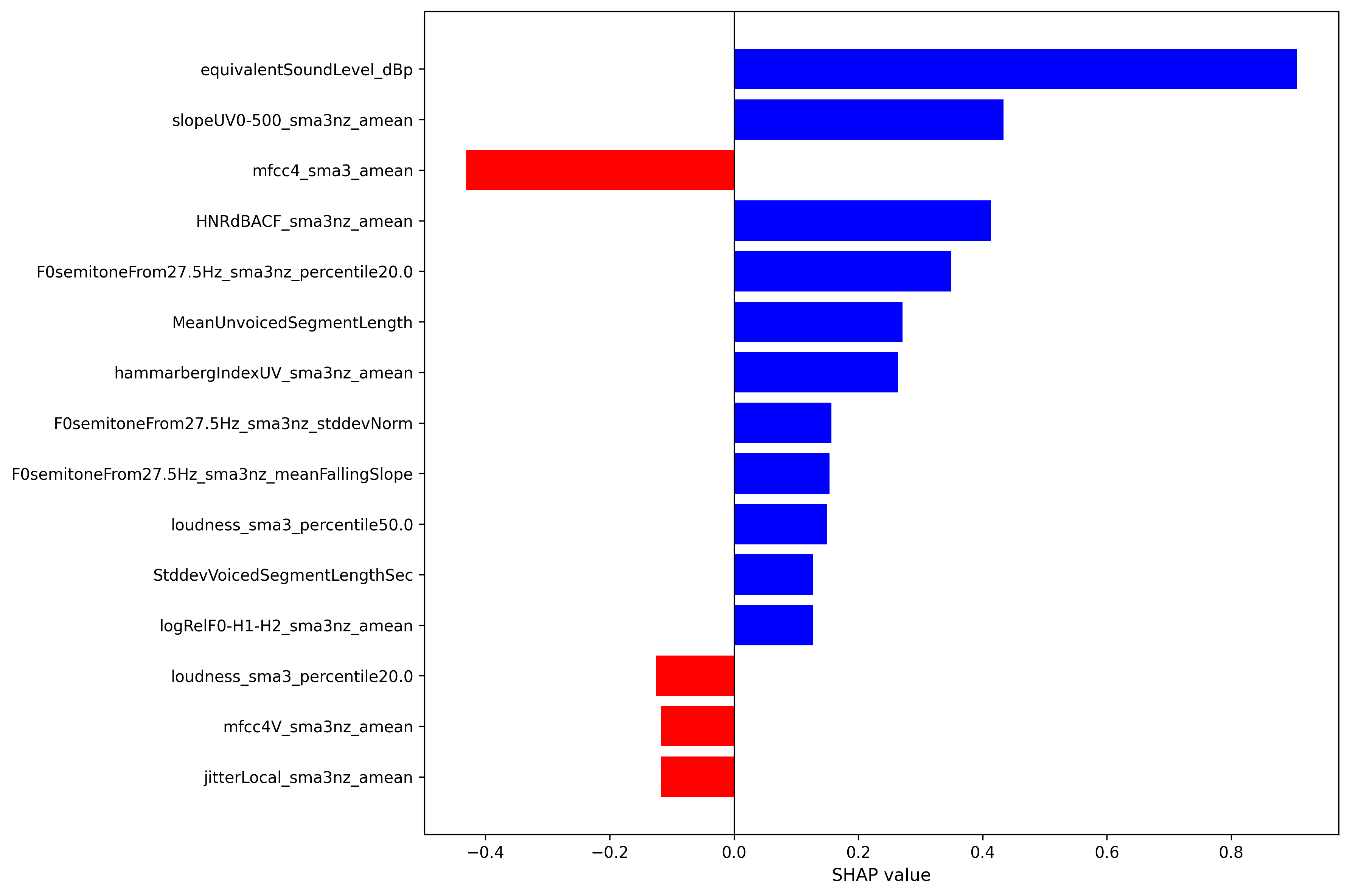}
  \caption{SHAP Feature Importance}
\end{subfigure}
\hfill
\begin{subfigure}{0.45\linewidth}
  \includegraphics[width=\linewidth]{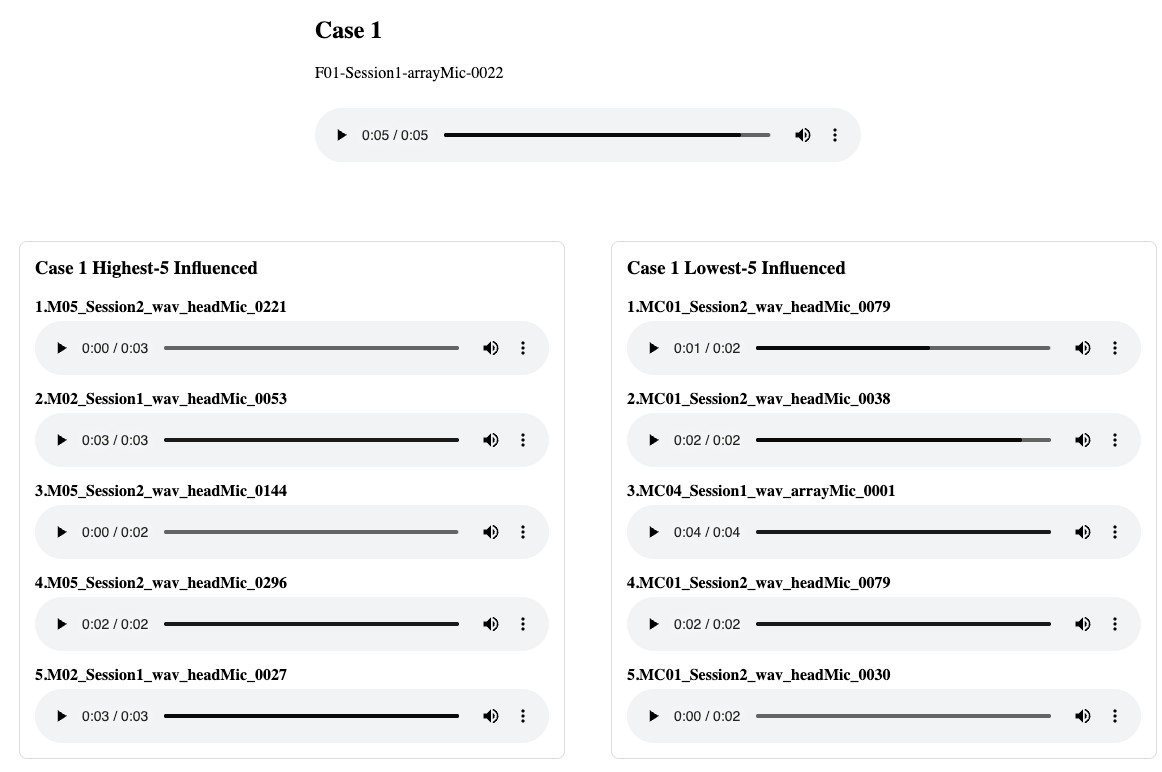}
  \caption{Influence-Based Explanation}
\end{subfigure}
\caption{Comparison of explanations generated by SHAP (left) and the proposed influence-based method (right) for a severe dysarthria case. 
}
\vspace{-10pt}
\label{fig:case_study}
\end{figure}

We qualitatively compare our proposed instance-based explanations with feature-attribution explanations for a severe dysarthria case in TORGO in Figure~\ref{fig:case_study}. 
As presented in Figure~\ref{fig:case_study}(a), SHAP~\cite{shap} produces importance score over 88 OpenSMILE acoustic descriptors 
(e.g., \textit{equivalentSoundLevel\_dBp}, \textit{mfcc4\_sma3\_amean}).
While these outputs are acoustically detailed, the descriptor names and scores are difficult to interpret in clinical terms (e.g., for assessment) and not directly support perceptual verification of whether the model's decision aligns with dysarthria markers used by clinicians.

In contrast, Figure~\ref{fig:case_study}(b) shows that our instance-based explanations method returns ranked training utterances with supportive and opposing influence (and the scores) for the same test case. Because the explanations are audio recordings, clinicians can verify their decisions through direct perceptual comparison between the supportive/competing samples with the reference cases and can inspect whether competing evidence arises from adjacent severity levels. Audio examples are provided on the demo page for perceptual test: \url{https://sites.google.com/view/infx-dys-samples}.

When extracting the top-5 most influential training utterances for multiple severe test cases, we observed that many test cases shared nearly identical top-ranked influential training examples. Manual inspection revealed that several of these high-influence training files were silent recordings (no words/sentences spoken). Notably, silence removal had already been applied during preprocessing, making it surprising that such near-silent segments still emerged among the most influential samples.
These files appear to originate from severe speakers, meaning the model may learn a spurious association between near-silence and the severe label. Importantly, the influence ranking surfaced these issues directly, highlighting the framework's practical utility for dataset auditing and its faithfulness in identifying training evidence that truly drives predictions---not only interpretable but also \emph{operationally useful}.

\section{Conclusion}

We proposed an influence-based instance-level explainability framework for dysarthria severity assessment that explains each prediction through training utterances that provide supporting and opposing evidence/influence, enabling perceptual verification via reference audio examples. Through controlled deletion experiments, we validated that estimated influence scores identify genuine prediction-relevant training samples: removing high-influence training samples consistently degrades performance, whereas removing low-influence samples can improve it. Empirical cross-severity analysis further reveals the model encodes severity as an ordered continuum with ordinal sensitivity reflected in influence decaying systematically with increasing label distance. 
Future work will extend this framework to larger and more diverse clinical populations and investigate robustness and cross-condition generalization under varying recording conditions and label granularities.

\newpage

\section{Acknowledgements}
This work was supported by the Engineering and Physical Sciences Research Council (EPSRC) through the National Edge AI Hub for Real Data: Edge Intelligence for Cyberdisturbances and Data Quality (EP/Y028813/1) and Responsible AI UK (EP/Y009800/1).

\section{Generative AI Tools Disclosure}
Generative artificial intelligence tools were used solely to assist with language editing and clarity of presentation. All research ideas, methodology, experiments, and interpretations were conceived and carried out by the authors, who take full responsibility for the originality, validity, and integrity of the work.

\bibliographystyle{IEEEtran}
\bibliography{mybib}

\end{document}